\documentclass[lettersize,journal]{IEEEtran}
\usepackage{amsmath,amsfonts}
\usepackage{algorithmic}
\usepackage{algorithm}
\usepackage{array}
\usepackage[caption=false,font=normalsize,labelfont=sf,textfont=sf]{subfig}
\usepackage{textcomp}
\usepackage{stfloats}
\usepackage{url}
\usepackage{verbatim}
\usepackage{graphicx}
\usepackage{cite}

\usepackage{amssymb}
\usepackage{multirow}
\usepackage{threeparttable}
\usepackage{arydshln}
\usepackage[pagebackref=true,breaklinks=true,letterpaper=true,colorlinks,bookmarks=false]{hyperref}

\begin{document}

\title{DiffX: Guide Your Layout to Cross-Modal Generative Modeling}

\author{Zeyu Wang,~\IEEEmembership{Student Member, IEEE}, Jingyu Lin, Yifei Qian, Yi Huang, Shicen Tian, Bosong Chai,\\~\IEEEmembership{Student Member, IEEE},
Juncan Deng,~\IEEEmembership{Student Member, IEEE}, Qu Yang, Lan Du,~\IEEEmembership{Senior Member, IEEE}, \\Cunjian Chen,~\IEEEmembership{Senior Member, IEEE}, Kejie Huang,~\IEEEmembership{Senior Member, IEEE}

\thanks{This work was supported in part by National Science Foundation of China (NSFC) under Grant NO. 62274142.}
\thanks{Zeyu Wang, Shicen Tian, Bosong Chai, Juncan Deng, and Kejie Huang are with Zhejiang University, Hangzhou, China.
Jingyu Lin, Lan Du, and Cunjian Chen are with Monash University, Melbourne, Australia.
Yifei Qian is with University of Nottingham, Nottingham, United Kingdom.
Yi Huang is with Peking University, Beijing, China.
Qu Yang is with Wuhan University, Wuhan, China.
(E-mail: \{wangzeyu2020, tianshicen, chaibosong, dengjuncan, huangkejie\}@zju.edu.cn, \{jingyu.lin, lan.du, cunjian.chen\}@monash.edu, Yifei.Qian@nottingham.ac.uk, yihuang@stu.pku.edu.cn, yangqu@whu.edu.cn)
}
\thanks{Zeyu Wang and Jingyu Lin contributed equally to this work.}
\thanks{\emph{Corresponding authors: Cunjian Chen and Kejie Huang}.}
}

\maketitle

\begin{abstract}
  Diffusion models have made significant strides in language-driven and layout-driven image generation.
  However, most diffusion models are limited to visible RGB image generation.
  In fact, human perception of the world is enriched by diverse viewpoints, such as chromatic contrast, thermal illumination, and depth information.
  In this paper, we introduce a novel diffusion model for general layout-guided cross-modal generation, called DiffX.
  Notably, our DiffX presents a compact and effective cross-modal generative modeling pipeline, which conducts diffusion and denoising processes in the modality-shared latent space.
  Moreover, we introduce the Joint-Modality Embedder (JME) to enhance the interaction between layout and text conditions by incorporating a gated attention mechanism.
  To facilitate the user-instructed training, we construct the cross-modal image datasets with detailed text captions by the Large-Multimodal Model (LMM) and our human-in-the-loop refinement.
  Through extensive experiments, our DiffX demonstrates robustness in cross-modal ``RGB+X'' image generation on FLIR, MFNet, and COME15K datasets, guided by various layout conditions.
  Meanwhile, it shows the strong potential for the adaptive generation of ``RGB+X+Y(+Z)'' images or more diverse modalities on FLIR, MFNet, COME15K, and MCXFace datasets.
  To our knowledge, DiffX is the first model for layout-guided cross-modal image generation.
  Our code and constructed cross-modal image datasets are available at \url{https://github.com/zeyuwang-zju/DiffX}.
\end{abstract}


\section{Introduction}

\IEEEPARstart{H}{uman} perception of the world is greatly enhanced by diverse modalities beyond the visible spectrum, such as thermal imaging and depth information.
Cross-modal visual understanding typically leverages the source format of ``RGB+X'', where X represents additional data like Thermal (T) or Depth (D) images.
It exhibits a more comprehensive visual representation compared with RGB-only understanding, especially in complex environments.
Thermal images offer significant advantages in detecting objects in low-light conditions, while depth images provide rich information of spatial structures.
At present, a significant limitation in cross-modal visual understanding is the scarcity of pixel-aligned RGB+X training data, as the collection and registration processes are both challenging and time-consuming.

A promising solution to address the limitation is to utilize generative models for cross-modal data augmentation.
Over the past decade, there have been significant advancements in image generation, driven by the improvement of deep generative models such as Variational AutoEncoders (VAEs) \cite{KingmaW13} and Generative Adversarial Networks (GANs) \cite{goodfellow2014generative}.
Prominently, the recent diffusion models \cite{ddpm_theory, ho2020denoising} like DALL-E \cite{ramesh2021zero, ramesh2022hierarchical}, Imagen \cite{saharia2022photorealistic}, and Stable Diffusion (SD) \cite{rombach2022high} have gained great popularity for their ability to generate high-fidelity images.
Moreover, the field has witnessed the rise of layout-to-image models based on various layout conditions, including bounding boxes, semantic maps, and keypoints \cite{xie2023boxdiff, li2023gligen, zhang2023adding, mou2023t2i, ma2024directed, 10154005, kang2024image}, demonstrating the reliable performance in user-instructed image generation.

\begin{figure}[!tbp]
  \centering
  \includegraphics[width=1\linewidth]{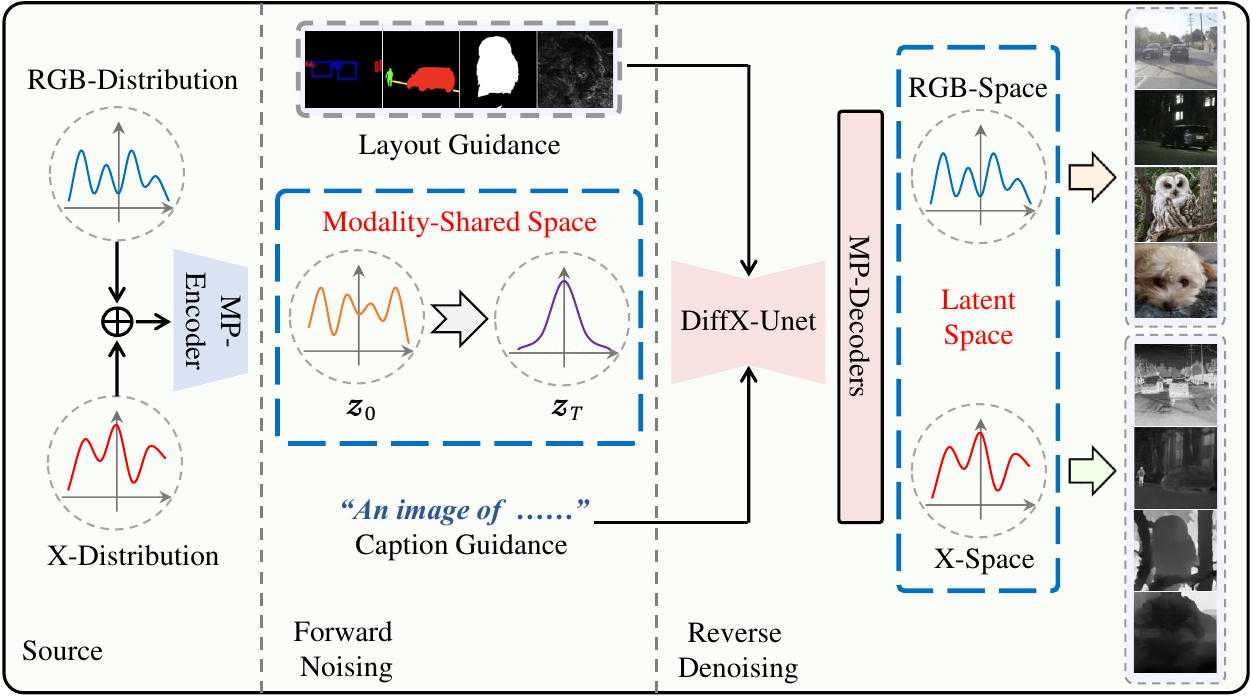}
  \caption{Illustration of our cross-modal generative pipeline in modality-shared latent space. 
  Here, the RGB+X modal encoding is employed for illustration, while the framework is capable of supporting additional modal inputs and outputs, such as RGB+X+Y and RGB+X+Y+Z.
  }
  \label{intro}
\end{figure}

However, current layout-to-image diffusion models primarily focus on the generation of RGB images in the visible spectrum.
When applied to cross-modal RGB+X generation, conventional diffusion models can only generate RGB and X images separately, resulting in misaligned and inconsistent image pairs.
It poses a significant challenge in data augmentation for cross-modal visual understanding.
Consequently, we wonder \textit{if we can generate the cross-modal images simultaneously under user instruction by an integrated model}?

\begin{figure*}[!tbp]
  \centering
  \includegraphics[width=1\linewidth]{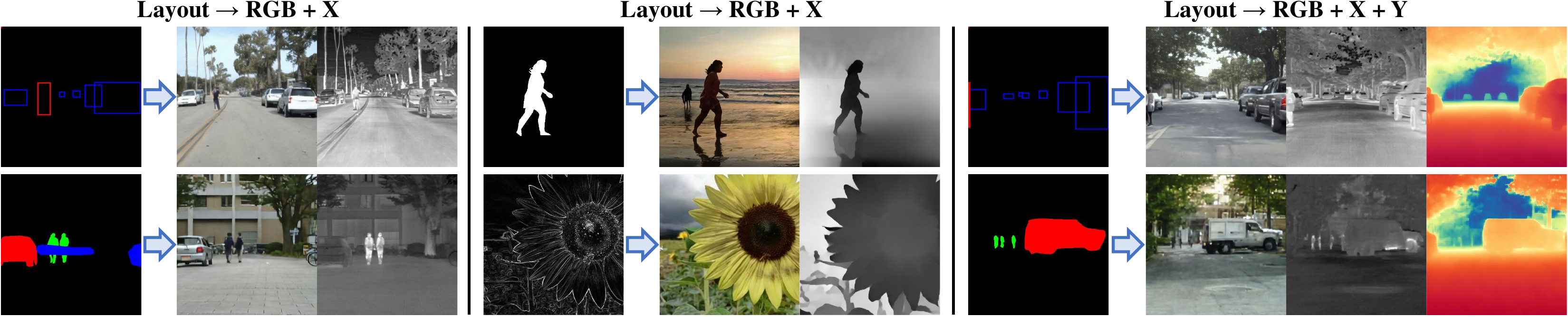}
  \caption{Illustration of the layout-guided cross-modal generation. The shown ``RGB+X+(Y)'' images are generated by our proposed DiffX model.
  In addition, our DiffX can also generate ``RGB+X+Y+Z'' images or more modalities.
  The X and Y images shown here can be Thermal (T) images or Depth (D) images.
  }
  \label{qualitative_head}
\end{figure*}

In this work, we present DiffX, a novel diffusion model designed for layout-guided cross-modal image generation utilizing a modality-shared latent diffusion mechanism, which is illustrated in Fig. \ref{intro}.
Through extensive experiments, our proposed DiffX model has shown its capability to generate RGB+X(+Y) images across diverse modalities guided by various layouts, as shown in Fig. \ref{qualitative_head}.

Our main contributions are summarized as follows:
\begin{itemize}
  \item We propose an effective cross-modal generative modeling pipeline, which performs the diffusion and denoising processes in the modality-shared latent space, facilitated by our Multi-Path Variational AutoEncoder (MP-VAE).
  \item We propose a Joint-Modality Embedder (JME) to establish the connection between layout and long text conditions via the gated cross-attention mechanism. 
  \item We introduce a human-in-the-loop method to construct cross-modal datasets with text captions, where we leverage the Large-Multimodal Model (LMM) for initial caption generation followed by expert manual corrections.
  \item Experiments demonstrate that DiffX can generate high-quality and well-aligned ``RGB+X'' images based on various layout conditions. Additionally, it shows strong adaptability to diverse ``RGB+X+Y(+Z)'' generation.
\end{itemize}

\section{Related Works}
  
\subsection{Cross-Modal Visual Understanding}
Cross-modal visual understanding normally adopts RGB+X input data for comprehensive visual representation.
Thermal images have shown great advantages in detecting objects in low-light conditions.
For instance, CFT \cite{qingyun2021cross} and ICAFusion \cite{shen2024icafusion} incorporate Transformer-based RGB+T feature fusion modules for multispectral object detection.
Additionally, SuperYOLO \cite{zhang2023superyolo} utilizes a super-resolution branch to improve small object detection in multispectral remote sensing imagery.
In semantic segmentation, BMDENet \cite{zhao2023bmdenet} introduces a bi-directional modality difference elimination module to mitigate heterogeneity between RGB+T images within the prototype space, further enhancing segmentation accuracy.
In Salient Object Detection (SOD), CCFENet \cite{CCFENet_TCSVT22} employs robust and accurate multi-modal expression encoding to integrate complementary multi-level features effectively.
Meanwhile, depth images exhibit valuable insights into spatial structures and 3D layouts.
For example, HidaNet \cite{wu2023hidanet} utilizes a granularity-based attention mechanism to enhance the advantages of RGB+D features for SOD task.
However, current cross-modal visual understanding is hindered by the lack of well-aligned RGB+X image pairs for training deep-learning models.

\subsection{Layout-to-Image Generation}
Based on Denoising Diffusion Probabilistic Model (DDPM) \cite{ddpm_theory, ho2020denoising}, diffusion models have undergone advancements in training and sampling techniques \cite{ddim, nichol2021improved, ho2022classifierfreediffusionguidance, yang2023eliminating, kang2024image, sun2024diffusion, deng2024vq4ditefficientposttrainingvector}. 
Recent studies have proposed layout-to-image diffusion models, which provide precise instruction for object generation.
For example, ControlNet \cite{zhang2023adding} and T2I-Adapter \cite{mou2023t2i} use the plug-in networks to learn the layout conditions.
LAW-Diffusion \cite{lawdiff} integrates a spatial parser and adaptive guidance, enabling complex scene generation.
BoxDiff \cite{xie2023boxdiff} directly incorporates spatial conditions into the training-free denoising process.
LayoutDiffusion \cite{layoutdiffusion} creates a cohesive layout for enhanced control in detailed object and global image generation.
Moreover, MIGC \cite{migc} focuses on enhanced position and attribute control in multi-instance generation. 
Additionally, InstanceDiffusion \cite{instancediffusion} focuses on precise instance-level control in high-quality image generation.
Interesting, Liang \emph{et al.} \cite{10121479} conducted text-to-image generation by creating an intermediate layout that connects the input text to the generated image.

\subsection{Multi-Modal Generation}
Currently, multi-modal generation has become a challenging task and attracted widespread research.
Notably, CoDi \cite{tang2024any} introduces a revolutionary approach enabling seamless generation across text, image, video, and audio by aligning prompt encoders and latent spaces of diffusion models.
MM-Diffusion \cite{ruan2023mm} integrates featuring two interconnected denoising autoencoders and a sequential multi-modal U-Net architecture for joint audio-video generation.
Moreover, MM-Interleaved \cite{tian2024mm} dynamically extracts information from multi-scale features for image-text generation.
MT-Diffusion \cite{chen2024diffusion} learns to generate various multi-modal data types with a multi-task loss.
4M \cite{mizrahi20244m} unifies multiple modalities' representation space for scalability, including text, images, geometric, and semantic conditions.
Despite these advancements, there is currently no generative model focused on layout-guided cross-modal ``RGB+X'' or ``RGB+X+Y(+Z)'' image generation.

\begin{figure*}[!tbp]
  \centering
  \includegraphics[width=1\linewidth]{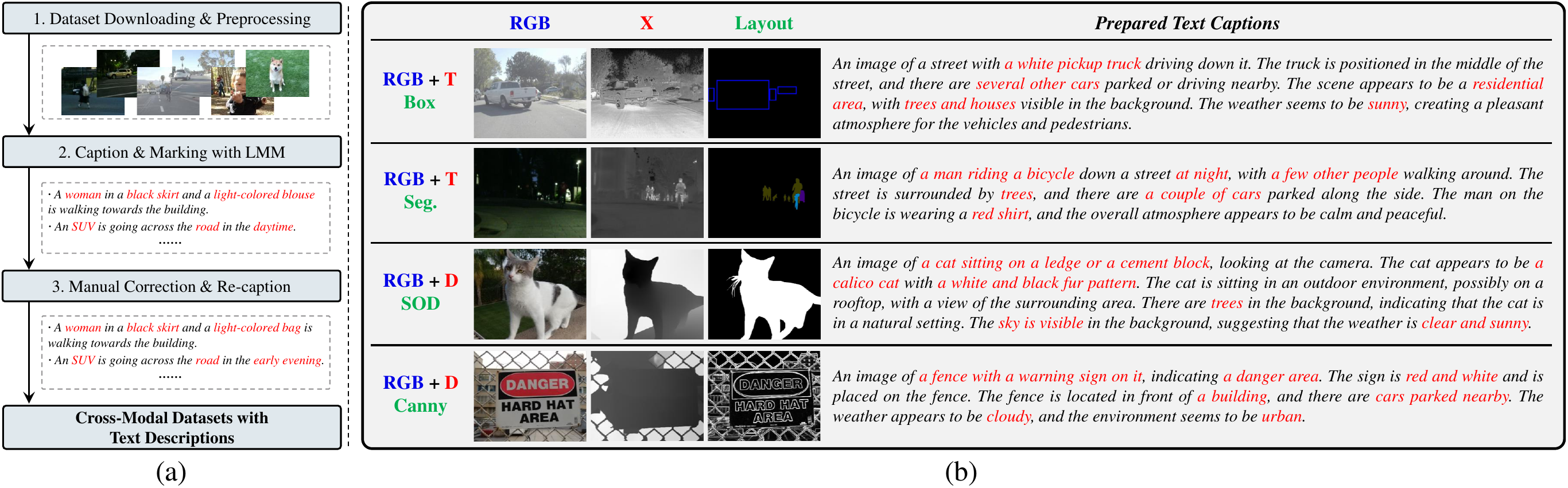}
  \caption{(a) The process of the human-in-the-loop method to construct the image captions. 
  (b) Examples of cross-modal images, labels, and prepared captions. 
  Within the layout types, Seg. denotes Semantic Segmentation map, and SOD denotes Salient Object Detection map.
  }
  \label{caption}
\end{figure*}

\section{Cross-Modal Dataset Construction}

Since our DiffX aims to generate cross-modal images under user instruction, it is crucial to construct datasets with accurate layout control and text descriptions for model training.
However, existing cross-modal datasets usually lack the specific modal images or detailed text captions.
In this section, we introduce the processes of constructing cross-modal datasets.

\textbf{Cross-Modal Image Construction}:
To construct the cross-modal image datasets, we first try to find public cross-modal datasets, such as the RGB+T or RGB+D datasets.
For tasks requiring the generation of RGB+T+D images, we leverage the pre-trained Marigold model \cite{ke2023repurposing} to derive the translated D images from the RGB images present in the RGB+T datasets.
Additionally, for some tasks, we aim to obtain Canny images as input conditions or output images, employing the Canny-Edge detection method \cite{4767851}.
The above approaches help us create datasets with diverse modalities for training our DiffX model with strong adaptability.

\textbf{LMM-Assisted Caption Generation}:
To obtain high-quality image captions for our cross-modal datasets, we employ the advanced LMM, namely LLaVA-v1.5-7b model \cite{liu2024visual}, to extract detailed descriptions of the RGB images by the given prompt.
We set the LLaVA prompt for caption generation on FLIR, MFNet, and COME15K datasets as: ``Based on the image, give me a description of this image, including its weather, environments, surroundings, transport conditions, trees, buildings, and others. Please use `An image of ...' to start the description.''
In addition, we set the prompt for human face caption generation on the MCXFace dataset as: ``Based on the image, help me describe this person's gender, appearance, hair, glasses, earrings, clothes, and other characteristics. Please use `An image of ...' to start the description.''

\textbf{Human-in-The-Loop Caption Correction}:
Despite the effectiveness of LLaVA in capturing visual content, it can sometimes generate inaccurate captions, especially in complex scenes.
Therefore, we conduct the human-in-the-loop processes to correct the generated image captions, as illustrated in Fig. \ref{caption} (a).
In cases where the generated captions exhibit minimal errors, we conduct manual corrections.
However, instances of significant errors or ambiguous descriptions necessitate further corrections. 
Under such conditions, we utilize GPT-4o \cite{achiam2023gpt} to rectify inaccurate information and provide additional context.
By combining the capability of the LLaVA model with the manual correction and assistance of GPT-4o, the captioning process ensures the accuracy in describing the visual context in the cross-modal datasets.
Examples of cross-modal data and prepared captions are shown in Fig. \ref{caption} (b).

\begin{figure*}[!tbp]
  \centering
  \includegraphics[width=1.0\linewidth]{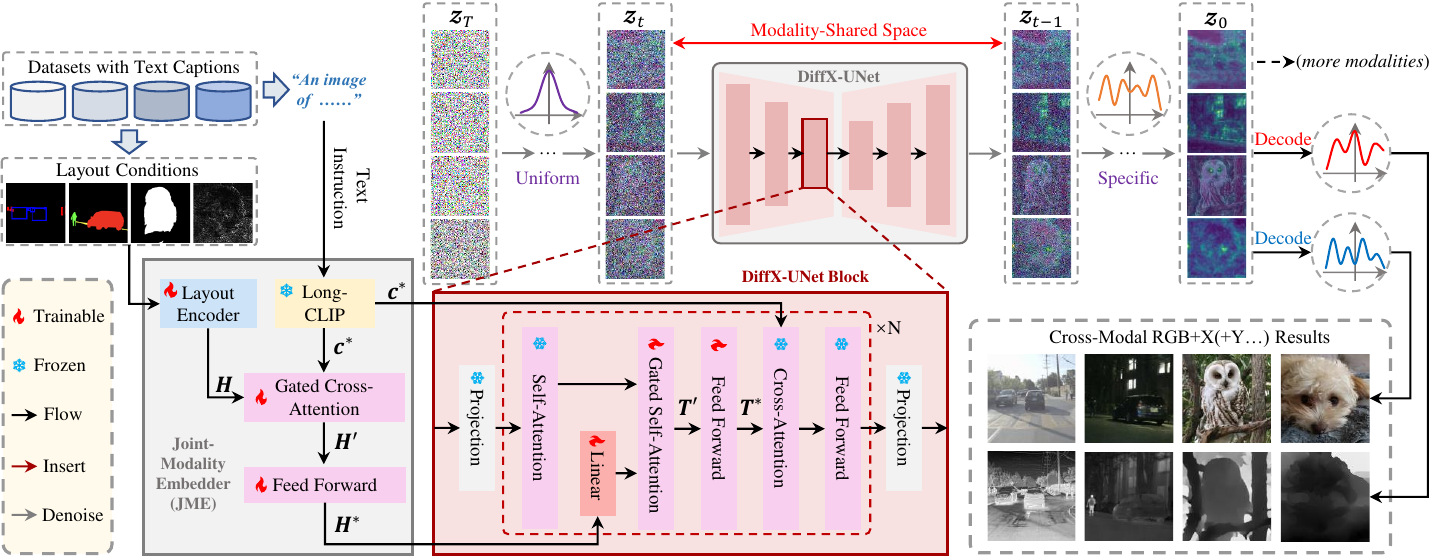}
  \caption{
    The workflow of our DiffX model for cross-modal generation.
    It performs the diffusion and denoising processes in the modality-shared space.
    Finally, the denoised feature $\boldsymbol{z}_0$ is decoded by the multi-path decoders into the cross-modal images in specific distributions.
  }
  \label{model}
\end{figure*}

\section{DiffX for Cross-Modal Generation}
The workflow of our DiffX model is shown in Fig. \ref{model}.
The noisy latent $\boldsymbol{z}_{T}$ is randomly initialized, which is adopted for generating $\boldsymbol{z}_{T-1}$, $\boldsymbol{z}_{T-2}$, · · ·, $\boldsymbol{z}_0$ by our DiffX-UNet.
Finally, the denoised feature $\boldsymbol{z}_0$ is decoded into the cross-modal images by the multi-path decoders of our MP-VAE.
The denoising process of the DiffX model involves embedding text captions and layout conditions for flexible user instruction.

In the following subsections, we first introduce the preliminaries on Latent Diffusion Model (LDM), and then we describe the workflow of our DiffX, including the modeling formulation, network structures, and optimization processes.

\subsection{Preliminaries on Latent Diffusion Model}
Recently, LDM and its successor SD \cite{rombach2022high} have pushed the boundaries of diffusion models by using a low-dimensional latent space.
Inside LDM, a prior VAE is employed to extract the latent representation $\boldsymbol{z}$ of the ground-truth image $\boldsymbol{I}$.
Subsequently, the basic denoising model $\boldsymbol{\epsilon}_{\boldsymbol{\vartheta } } $, normally implemented as a UNet \cite{ronneberger2015u} with residual \cite{he2016deep} and self-attention \cite{c:22} blocks, is adopted to obtain the noise $\boldsymbol{\epsilon}$ in the noisy latent $\boldsymbol{z}_t$ at time step $t$.
The training objective of condition-guided SD model can be represented as:
\begin{equation}
  \min _{\boldsymbol{\vartheta}} \mathcal{L}_{\text{LDM}}=\mathbb{E}_{\boldsymbol{z}_t, \boldsymbol{\epsilon} \sim \mathcal{N}(\mathbf{0}, \mathbf{I}), t, \boldsymbol{c}}\left[\left\|\boldsymbol{\epsilon}-\boldsymbol{\epsilon}_{\boldsymbol{\vartheta}}\left(\boldsymbol{z}_t, t, \boldsymbol{c}\right)\right\|_2^2\right],
\end{equation}
where $\boldsymbol{c}$ denotes the conditional input to the denoising UNet, which can be a text embedding generated by a text encoder like CLIP \cite{radford2021learning} in text-to-image generation.

Following the training process, the denoising model $\boldsymbol{\epsilon}_{\boldsymbol{\vartheta } } $ gradually generates $\boldsymbol{z}_{T-1}$, $\boldsymbol{z}_{T-2}$, · · ·, $\boldsymbol{z}_0$ from a randomly initialized noisy latent $\boldsymbol{z}_T$.
Finally, the decoder of VAE is employed to generate the target image $\boldsymbol{I}'$ based on $\boldsymbol{z}_0$.

\subsection{Cross-Modal Modeling Formulation}
As illustrated above, we propose an effective cross-modal generative modeling pipeline, which conducts the diffusion and denoising processes in the modality-shared latent space.
In our DiffX model, we first obtain the joint-modality distribution by mapping the cross-modal images into the shared latent space. 
Specifically, during the forward process, the noising transition is conditioned on the joint modality $\mathbf{M}=\{\boldsymbol{m}_1, \boldsymbol{m}_2, ..., \boldsymbol{m}_N\}$, where the modalities are independent of each other:
\begin{equation}
    q(\boldsymbol{z}_t,\mathbf{M}|\boldsymbol{z}_{t-1})=q(\boldsymbol{z}_t|\boldsymbol{z}_{t-1},\mathbf{M})\boldsymbol{\Pi}_{i=1}^{N}q_i(\boldsymbol{m}_i).
    \label{q}
\end{equation}

Meanwhile, the reverse process can be regarded as the converse of the forward distributions, resulting in a joint distribution $p_{\boldsymbol{\theta}}(\boldsymbol{z}_{t-1},\mathbf{M}|\boldsymbol{z}_{t})$ at time step $t$: 
\begin{equation}
    p_{\boldsymbol{\theta}}(\boldsymbol{z}_{t-1},\mathbf{M}|\boldsymbol{z}_{t})=p_{\boldsymbol{\theta}}(\boldsymbol{z}_{t-1}|\boldsymbol{z}_{t})\boldsymbol{\Pi}_{i=1}^{N}p_{\boldsymbol{\theta}}(\boldsymbol{m}_{i}|\boldsymbol{z}_{t}),
    \label{p}
\end{equation}
where $\boldsymbol{\theta}$ denotes the parameters of the DiffX model.
In the above equations, we assume $\boldsymbol{z}_{t-1}$ and $\mathbf{M}$ are conditionally independent given $\boldsymbol{z}_{t}$.

\subsection{Multi-Path Variational AutoEncoder}

\begin{figure}[!tbp]
  \centering
  \includegraphics[width=1\linewidth]{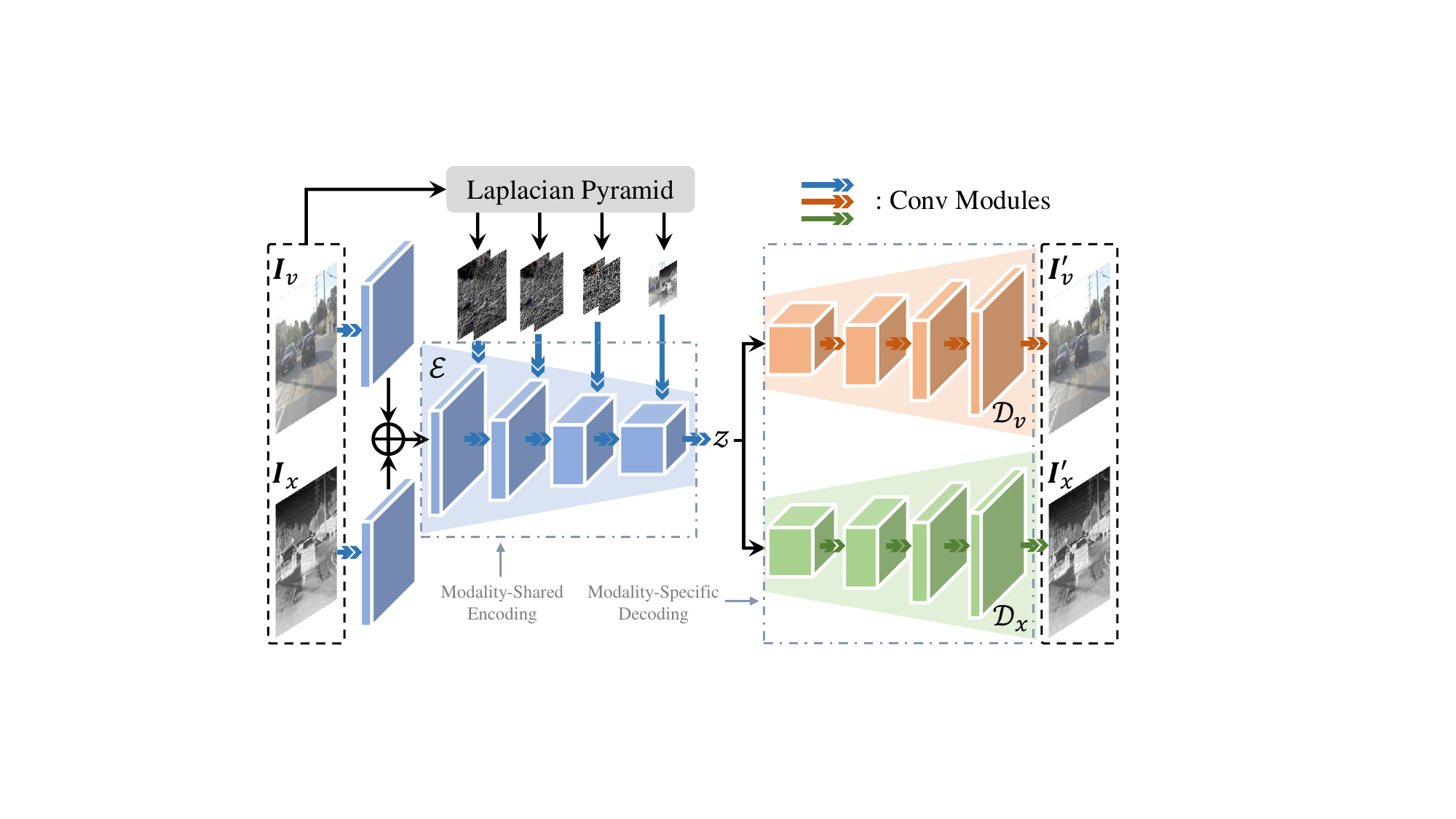}
  \caption{Workflow of our Multi-Path Variational AutoEncoder (MP-VAE).
  Here, the RGB+X modal encoding is employed for illustration, while it also supports additional modal encoding, such as RGB+X+Y and RGB+X+Y+Z.
  }
  \label{MP-VAE}
\end{figure}

As illustrated in Fig. \ref{MP-VAE}, the proposed MP-VAE is critical for our cross-modal generative task.
In contrast to the conventional VAE used in LDM, our MP-VAE employs the input and output form of ``RGB+X'' or more modalities.
Here, we take the bi-branch RGB+X modal encoding for illustration.
In detail, the MP-VAE utilizes a single encoder $\mathcal{E}$ to encode the input \{$\boldsymbol{I}_v, \boldsymbol{I}_x$\} (denoted as $\{\boldsymbol{m}_1, \boldsymbol{m}_2\}$ in Eq. (\ref{q})) into the modality-shared latent representation $z$.
The input images are processed through the separate convolutional layers, followed by element-wise addition before being fed to $\mathcal{E}$. 
Then, it incorporates parallel decoders $\mathcal{D}_v$ and $\mathcal{D}_x$ to generate the corresponding output \{$\boldsymbol{I}_v', \boldsymbol{I}_x'$\}.

In addition, we have observed that the modality ``X'' (usually Thermal or Depth images) exhibits distinct contour edges and prominent object positions.
To enhance the reconstruction ability of the MP-VAE, we apply a Laplacian Pyramid (LP) \cite{burt1987laplacian} to extract the high-frequency information from the input cross-modal image pairs.
We adopt the Laplacian pyramid feature extraction strategy in the LPTN algorithm \cite{liang2021high}.
Subsequently, the extracted features are embedded into the multi-scale layers of the encoder through cascaded convolutional layers, enabling a comprehensive enhancement at various frequencies.

During the inference phase of DiffX, the parallel decoders from the pre-trained MP-VAE are employed to generate the cross-modal ``RGB+X(+Y+Z)'' images or more diverse modalities based on the denoised latent feature $\boldsymbol{z}_0$.

\subsection{Joint-Modality Embedder}
Our proposed DiffX aims to guide the layout conditions with the long text captions for cross-modal generation.
To establish a comprehensive connection between the layout ($\boldsymbol{b}$ or $\boldsymbol{s}$) and the text ($\boldsymbol{c}$), we propose the Joint-Modality Embedder (JME, denoted as $\mathcal{J}_\omega $) to obtain the conditional features:
\begin{equation}
  \boldsymbol{H}^{*}, \boldsymbol{c}^{*} = \mathcal{J}_\omega(\boldsymbol{b}/\boldsymbol{s}, \boldsymbol{c}),
\end{equation}
where $\boldsymbol{H}^{*}$ is the text-aware layout feature, and $\boldsymbol{c}^{*}$ is the caption feature.
In detail, $\mathcal{J}_\omega$ consists of the layout encoder, the Long-CLIP text encoder \cite{zhang2024long}, a gated cross-attention layer, and a Feed Forward (FF) layer.
The FF layer is a Multi-Layer Perception (MLP) \cite{MLP} with middle-dimension expansion.

\subsubsection{Layout Condition Embedding}
Our DiffX can guide various types of layout conditions for cross-modal generation.
Different from ControlNet \cite{zhang2023adding}, our DiffX embeds the layout conditions into the feature of fixed dimensions and feed them into the gated self-attention layers.

For the box-based layout, boxes $\boldsymbol{b}_i$ and labels $\boldsymbol{l}_i$ are embedded via Fourier mapping $\mathcal{F} (\cdot)$ \cite{tancik2020fourier} and CLIP text encoder $f_{\text{clip}}(\cdot)$, respectively.
Then, an MLP with parameters $\phi$ is used to encode them into grounding tokens:
\begin{equation}
  \boldsymbol{h}_i = \text{MLP}_{\phi }([\mathcal{F}(\boldsymbol{b}_i), f_{\text{clip}}(\boldsymbol{l}_i)]); \boldsymbol{H} = [\boldsymbol{h}_1, ..., \boldsymbol{h}_n],
\end{equation}
where $[\cdot]$ denotes the feature concatenation.
The representation $\boldsymbol{H}$, which contains a total of $n$ embedded box features, is adopted as the layout conditional feature.

For the semantic mask layout represented by $\boldsymbol{s}$, we utilize the pre-trained ConvNeXt$_{\psi}$ model \cite{liu2022convnet} to extract the in-depth semantic feature.
Subsequently, the position embedding $\boldsymbol{p}$ is added, and an MLP with parameters $\varphi$ is used to generate the corresponding layout feature:
\begin{equation}
  \boldsymbol{H} = \text{MLP}_{\varphi }(\mathcal{R} (\text{ConvNeXt}_{\psi}(\boldsymbol{s})) + \boldsymbol{p}),
\end{equation}
where $\mathcal{R}$ denotes the process of reshaping and flattening the semantic feature into tokens.

\subsubsection{Long-CLIP Caption Embedding}
Text conditions $\boldsymbol{c}$ remain equally significant in cross-modal generation as they provide detailed image descriptions, including various aspects such as weather, environments, and transportation.
Specifically, our task necessitates long text captions to describe the small objects and complex backgrounds in the cross-modal scene, as shown in Fig. \ref{caption} (b).
In our DiffX, we utilize the advanced Long-CLIP as the text encoder, which represents a significant improvement over its predecessor, CLIP, due to its ability to support text inputs of up to 248 tokens.
In detail, each caption $\boldsymbol{c}$ is embedded into a sequence of text embeddings, which can be formulated as $\boldsymbol{c}^{*} = f_{\text{L-clip}}(\boldsymbol{c})$.

\subsubsection{Gated Cross-Attention}
Inside our $\mathcal{J}_\omega $, the gated Cross-Attention (CA) layer is implemented to fuse the layout and text conditional features, which is formulated as follows:
\begin{equation}
  \text{CA}(\boldsymbol{H}, \boldsymbol{c}^{*}) = \text{softmax}\left(\frac{W_q(\boldsymbol{H})\cdot W_k(\boldsymbol{c}^{*})^T}{\sqrt{d_k}}\right)\cdot W_v(\boldsymbol{c}^{*}),
\end{equation}
where $W_q$, $W_k$, and $W_v$ denote the linear transformations, and $d_k$ denotes the dimension of $W_k(\boldsymbol{c}^{*})$.
Here, the CA layer derives query from layout condition $\boldsymbol{H}$, while deriving key and value from text condition $\boldsymbol{c}^{*}$.

To obtain the text-aware layout conditional feature $\boldsymbol{H}^{*}$, a gated CA layer and an FF layer are sequentially implemented in $\mathcal{J}_\omega$, which can be formulated as:
\begin{equation}
  \boldsymbol{H}' = \boldsymbol{H} + \lambda \cdot \text{tanh}(\gamma_1) \cdot \text{CA}(\boldsymbol{H}, \boldsymbol{c}^{*}),
\end{equation}
\begin{equation}
  \boldsymbol{H}^{*} = \boldsymbol{H}' + \lambda \cdot \text{tanh}(\gamma_2) \cdot \text{FF}(\boldsymbol{H}'),
\end{equation}
where $\boldsymbol{H}'$ is the intermediate layout feature between the gated CA layer and FF layer.
The hyperparameter $\lambda$ plays a crucial role in balancing feature quality and controllability.
Additionally, $\gamma_1$ and $\gamma_2$ are two adaptive learnable scalars of the tanh activation.

\subsection{DiffX-UNet}

As illustrated in Fig. \ref{model}, DiffX-UNet serves as the basic model for the denoising process in the modality-shared latent space.
It is composed of residual projection blocks and Spatial-Transformer blocks.
The Spatial-Transformer block is the critical component of the DiffX-UNet, consisting of cascaded attention-based layers and FF layers.
To enhance the training efficiency, pre-trained model weights from SD-v1-4 \cite{Rombach_2022_CVPR} are initialized for the frozen layers.

The joint-modality feature $\boldsymbol{H}^{*}$ embedded by $\mathcal{J}_\omega$ is fed into the gated Self-Attention (SA) layer, which acts as an adapter module to capture the cross-modal relationships.
Similar to the gated CA layer, an FF layer is integrated after each gated SA layer to build the long-range interaction:
\begin{equation}
  \boldsymbol{T}' = \boldsymbol{T} + \mu \cdot \text{tanh}(\delta_1 ) \cdot \mathcal{T}  (\text{SA}([\boldsymbol{T}, f_\text{linear}(\boldsymbol{H}^{*})])),
\end{equation}
\begin{equation}
  \boldsymbol{T}^{*} = \boldsymbol{T}' + \mu \cdot \text{tanh}(\delta_2) \cdot \text{FF}(\boldsymbol{T}'),
\end{equation}
where $\mathcal{T}(\cdot)$ is a token selection operation that selects the visual token positions of $\boldsymbol{T}$.
The hyperparameter $\mu$ adjusts the weights of the SA mechanism, whereas $\delta_1$ and $\delta_2$ are also two learnable scalars.
For the trainable FF layers, we also add the scaling factors with the tanh activation.

Compared with the conventional Transformer blocks in SD, our Spatial-Transformer blocks integrate the iterative gated SA layers to adaptively embed the joint-modality layout feature $\boldsymbol{H}^{*}$.
Moreover, the text condition $\boldsymbol{c}^{*}$ embedded by Long-CLIP further enhances the long caption awareness.

\begin{algorithm}[!tbp]
	\caption{Optimization process of DiffX.}
	\renewcommand{\algorithmicrequire}{\textbf{STEP1:}}
	\renewcommand{\algorithmicensure}{\textbf{STEP2:}}
	\begin{algorithmic}[1]
		\REQUIRE Training MP-VAE.
		\STATE \textbf{While} training:
		\STATE \hspace{0.3cm} Forward:  $\{\boldsymbol{I}_v', \boldsymbol{I}_x'\}=\text{MP-VAE}(\{\boldsymbol{I}_v, \boldsymbol{I}_x\})$;
		\STATE \hspace{0.3cm} Minimize:  $\tilde{\mathcal{L}}_\text{MP}$ formulated in Eq. (\ref{DP});
    \STATE \hspace{0.3cm} Update: MP-VAE;
    \STATE \textbf{End While}.

    \hspace{0.3cm}

    \ENSURE Training DiffX (DiffX-UNet denoted as $\boldsymbol{\epsilon}_{\boldsymbol{\vartheta}}$).
    \STATE Define: $\alpha_1$, $\alpha_2$, ..., $\alpha_T$ (derived from $\beta_1$, $\beta_2$, ..., $\beta_T$);
		\STATE \textbf{While} training:
    \STATE \hspace{0.3cm} Forward:  $\boldsymbol{z}_{0}= \text{DP-Encoder} (\{\boldsymbol{I}_v, \boldsymbol{I}_x\})$;
    \STATE \hspace{0.3cm} Forward:  $\boldsymbol{c}^{*}$ and $\boldsymbol{H}^{*}$ based on $\boldsymbol{c}$ and $\boldsymbol{b}/\boldsymbol{m}$;
    \STATE \hspace{0.3cm} Sample: $t \sim \text{Uniform}(\{1, ..., T\}) $, $\boldsymbol{\epsilon} \sim \mathcal{N}(\mathbf{0}, \mathbf{I})$;
    \STATE \hspace{0.3cm} Calculate: $\boldsymbol{z}_t = \sqrt{\overline{\alpha}_t }\boldsymbol{z}_0 + \sqrt{1-\overline{\alpha}_t } \boldsymbol{\epsilon}$;
    \STATE \hspace{0.3cm} Minimize: $\mathbb{E}_{\boldsymbol{z}_0, \boldsymbol{\epsilon}, t}\left[\|\boldsymbol{\epsilon}-\boldsymbol{\epsilon}_{\boldsymbol{\vartheta}}\left(\boldsymbol{z}_t, t, \boldsymbol{c}^{*}, \boldsymbol{H}^{*}\right)\|_2^2\right]$;
    \STATE \hspace{0.3cm} Update: Trainable modules of DiffX;
    \STATE \textbf{End While}.
	\end{algorithmic}
	\label{optimization}
\end{algorithm}

\begin{figure*}[!tbp]
  \centering
  \includegraphics[width=1\linewidth]{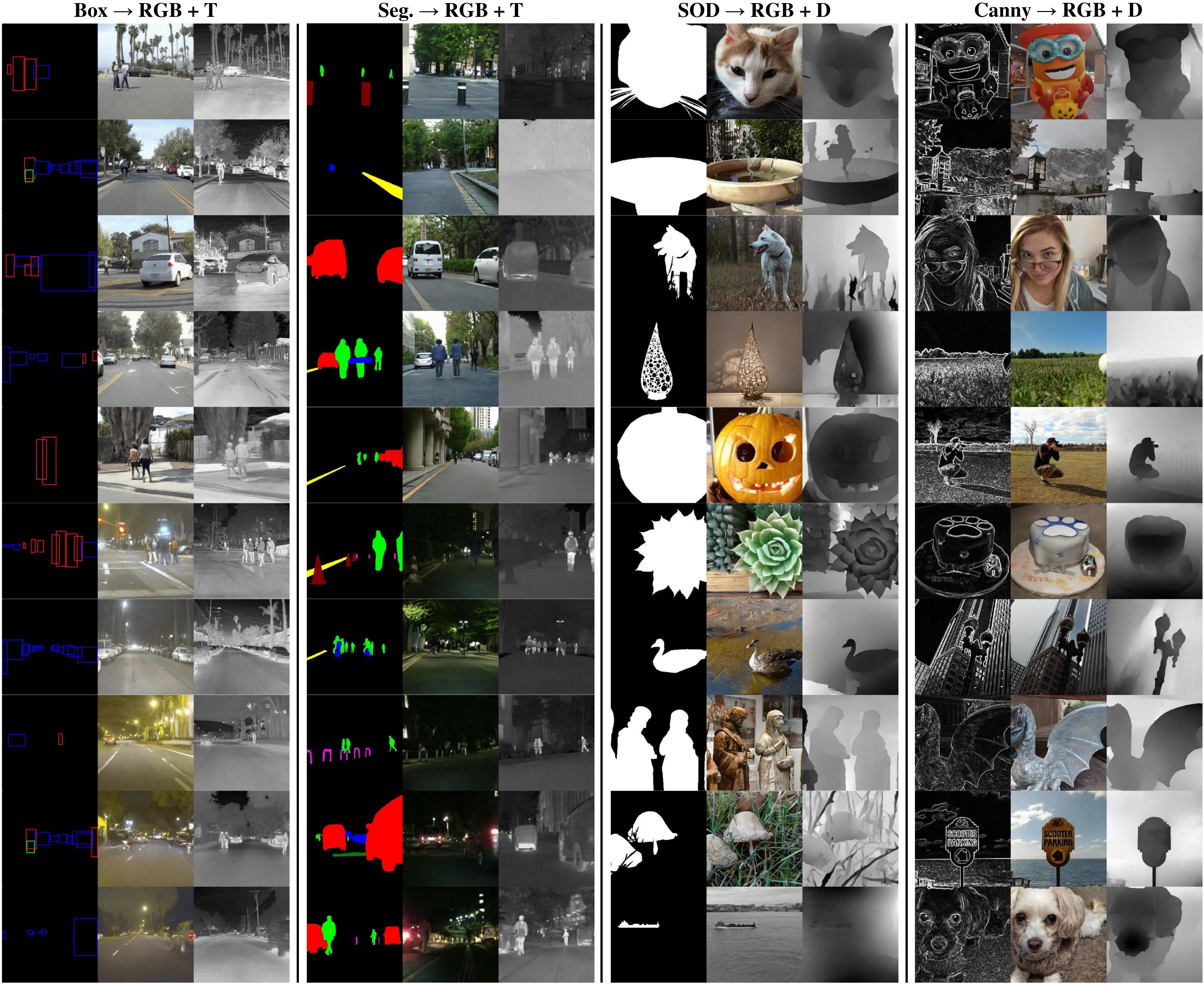}
  \caption{Qualitative results of cross-modal ``RGB+X'' generation under various layout guidance by our DiffX. The synthesis resolution is fixed to 512 $\times$ 512.
  }
  \label{qualitative}
\end{figure*}

\subsection{Training \& Inference}
Here, we illustrate our training and inference strategies.
\subsubsection{Training}
The optimization processes and details of our proposed DiffX model are shown in Algorithm \ref{optimization}.

Firstly, the training of MP-VAE is critical prior to the training of DiffX for cross-modal generation, where we encode the cross-modal images into the modality-shared latent space $\boldsymbol{z}$ and conduct end-to-end reconstruction. 
The training objective of MP-VAE includes the Mean Squared Error (MSE) loss and the perceptual loss \cite{johnson2016perceptual} fitted with Kullback-Leibler (KL) divergence, which can be formulated as:
\begin{equation}
  \tilde{\mathcal{L}}_\text{MP} \triangleq \mathcal{L}_{\text{mse}}(\mathbf{M}, \widehat{\mathbf{M}} ) + \boldsymbol{\Sigma}_{i=1}^N \text{KL} \left(q_i(\boldsymbol{m}_i)\|p_{\boldsymbol{\theta}}(\boldsymbol{m}_i|\boldsymbol{z})\right).
  \label{DP}
\end{equation}

Secondly, the training objective of our DiffX model, namely the latent denoising function, is formulated as:
\begin{equation}
  \tilde{\mathcal{L}}_{\text{DiffX}}\triangleq\mathbb{E}_{q}[\boldsymbol{\Sigma}_{t>1}\left\|\boldsymbol{\epsilon} - \boldsymbol{\epsilon}_{\boldsymbol{\vartheta }}(\boldsymbol{z}_{t},t,\boldsymbol{c}^{*}, \boldsymbol{H}^{*})\right\|_{2}^{2}], 
\end{equation}
where $t \sim \text{Uniform}(\{1, ..., T\})$ is sampled in each iteration. 
Then, the latent $\boldsymbol{z}_t\sim q(\boldsymbol{z}_t|\boldsymbol{z}_0,\mathbf{M})$ is initialized, which is fed into DiffX to predict the forward noise. 
Finally, we optimize the distance between the predicted noise and target noise.

\subsubsection{Inference}
The difference in the sampling steps between our DiffX and conventional LDM is the multi-branch decoding by our MP-VAE, which is formulated in Eq. (\ref{q}).
The main purpose is to estimate cross-modal images from the denoised features of the corresponding modality-shared latent distributions based on the well-trained DiffX model.

\begin{figure*}[!tbp]
  \centering
  \includegraphics[width=1\linewidth]{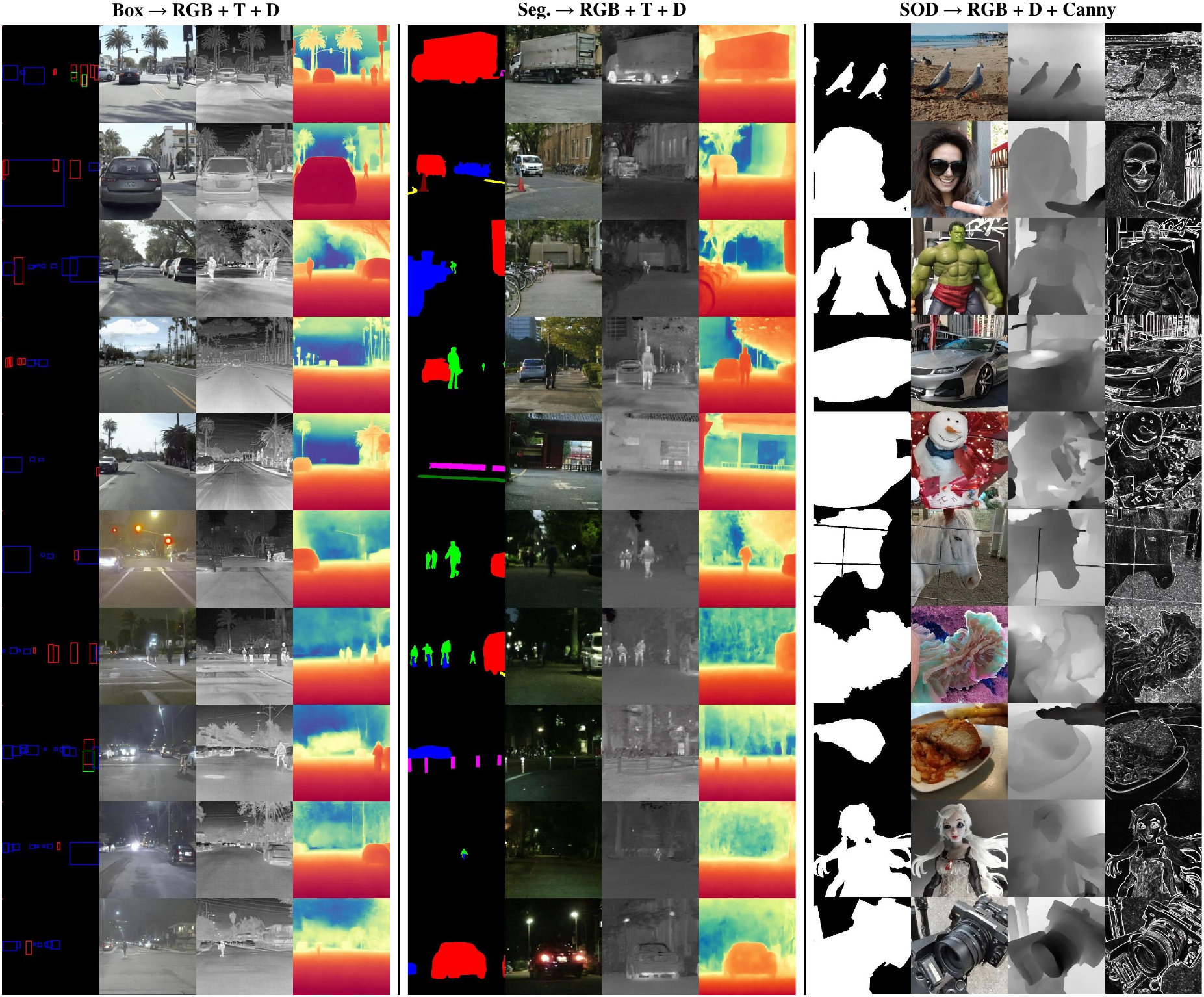}
  \caption{Qualitative results of layout-guided cross-modal ``RGB+X+Y'' generation by our DiffX. The synthesis resolution is fixed to 512 $\times$ 512.
  Note that there are two types of Depth (D) maps: colored and gray, both of which contain depth information.
    }
  \label{triple}
\end{figure*}

\begin{figure}[!tbp]
  \centering
  \includegraphics[width=1\linewidth]{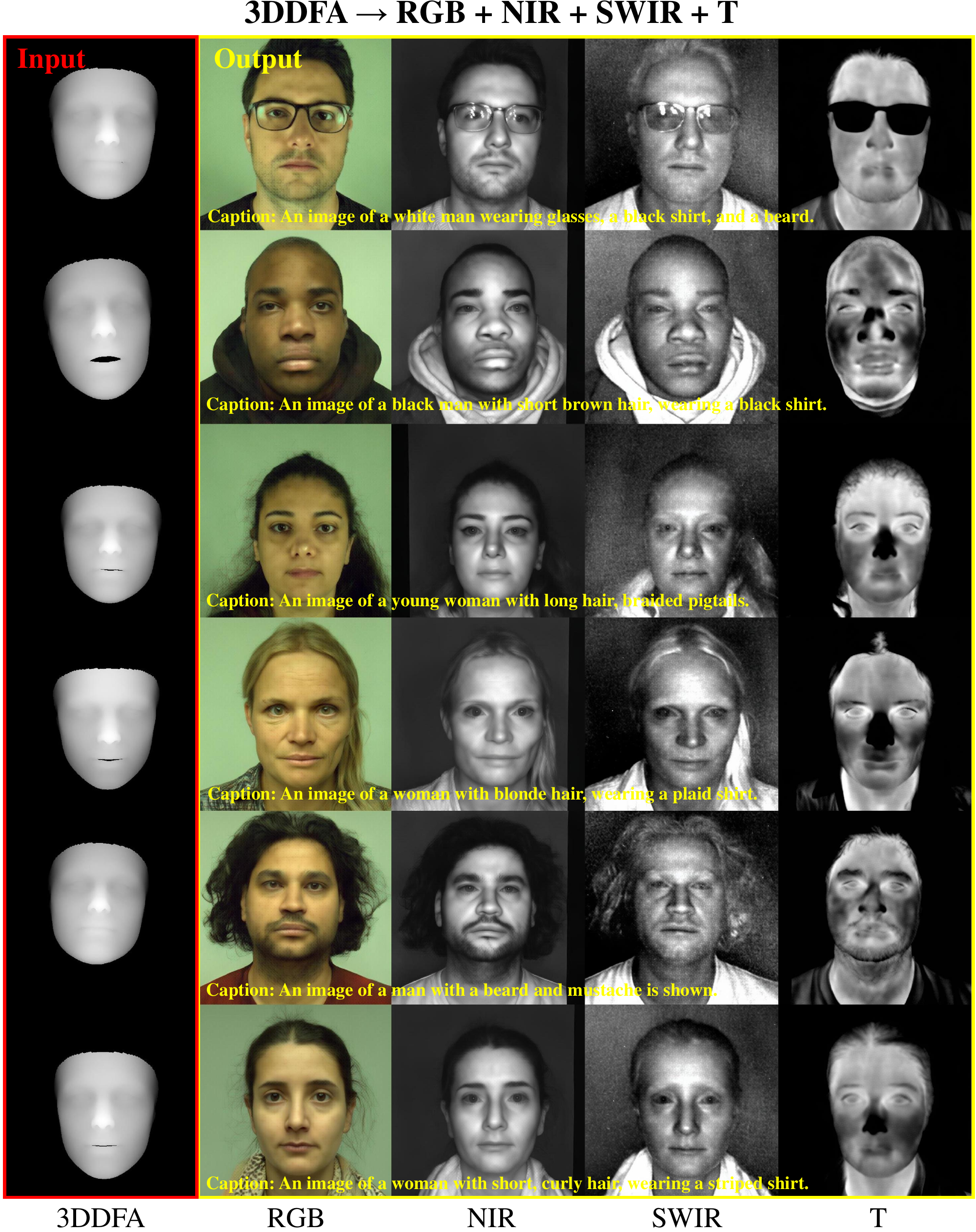}
  \caption{
      Qualitative results of ``3DDFA$\rightarrow$RGB+NIR+SWIR+T'' multispectral face image generation by our DiffX on the MCXFace dataset. 
    }
  \label{forth}
\end{figure}

\section{Experiments}

\subsection{Implementation Details}
We implement our DiffX model with PyTorch 1.13 and CUDA 12.2 on four NVIDIA RTX A6000 GPUs with an Intel(R) Xeon(R) Gold 5218 CPU.
The learning rate is set to 5$e-$5 with 10,000 warm-up steps in a constant scheduler.
During the training phase, we set the probability of 0.5 to randomly drop the caption condition, where $\boldsymbol{c}^{*}$ will be an all-zero text embedding.
The batch sizes are set to 4 $\times$ 2 and 4 $\times$ 8 for training our MP-VAE and DiffX-UNet, respectively.
We fine-tune the MP-VAE using the pre-trained VAE from SD model until convergence.
The sampling resolution of cross-modal images is fixed to 512 $\times$ 512 for inference.

\subsection{Cross-Modal Datasets}
In this work, we conduct extensive experiments on four datasets for eight cross-modal generative tasks.
In detail, we adopt the clean-version FLIR dataset \cite{FLIR2019, zhang2020multispectral} for ``Box $\rightarrow$ RGB+T(+D)'' task and the MFNet dataset \cite{ha2017mfnet} for ``Segmentation (Seg.) $\rightarrow$ RGB+T(+D)'' task.
Since the two datasets only contain RGB+T image pairs, we utilize the pre-trained Marigold \cite{ke2023repurposing} to obtain the translated colored D images.
In addition, the COME15K dataset \cite{cascaded_rgbd_sod} is adopted for ``SOD $\rightarrow$ RGB+D'', ``Canny $\rightarrow$ RGB+D'', and ``SOD $\rightarrow$ RGB+D+Canny'' tasks.
Meanwhile, we utilize MCXFace dataset \cite{george2022prepended} for ``3DDFA $\rightarrow$ RGB+NIR+SWIR+T'' task on multispectral human face images, where the 3DDFA \cite{zhu2017face} is the 3D face layout, while NIR and SWIR denote the near-infrared images and short-wavelength infrared images, respectively.

The official dataset splits in FLIR, MFNet, and COME15K datasets are adopted for the training and testing phases in our experiments.
For the MCXFace dataset, we split the human face images from 21 people for testing, and the remaining images are adopted for training.
All qualitative results shown in the work are obtained from the test sets of the datasets.

\subsection{Evaluation Metrics}
During the evaluation phase, we utilize Learned Perceptual Image Patch Similarity (LPIPS)  \cite{zhang2018unreasonable} and Fréchet Inception Distance (FID)  \cite{heusel2017gans} to evaluate the generated cross-modal image quality.
Meanwhile, we adopt the pre-trained models in downstream vision tasks to evaluate the correspondence to the original layouts.
We use CFT\cite{qingyun2021cross} to test Average Precision (AP) and AP$_{50}$ on Box $\rightarrow$ RGB+T task, EGFNet \cite{zhou2022edge} to test mean Intersection over Union (IoU) and mean Accuracy (Acc) on Seg. $\rightarrow$ RGB+T task, and BBSNet \cite{fan2020bbsnet} to test S-measure ($S_{\alpha}$) \cite{fan2017structure} and Mean Absolute Error ($M$) scores on Canny $\rightarrow$ RGB+D task.
These scores aim to evaluate the consistency of the generated cross-modal images and the layout conditions.
Additionally, we employ the Structural Similarity (SSIM) score \cite{wang2004image} to evaluate the consistency of generated and ground-truth T or D images, denoted as SSIM$_x$.
In the ablation study, we use Peak Signal-To-Noise Ratio (PSNR) and SSIM to evaluate the cross-modal reconstruction by MP-VAE.

\subsection{Experimental Results and Comparison}

\subsubsection{Unified Cross-Modal Generation}
As shown in Fig. \ref{qualitative}, we can see that DiffX is a unified framework for cross-modal ``RGB+X'' generation under various layout conditions.
The RGB+T generation benefits from precise control over object positions and shapes.
Meanwhile, it excels at RGB+D generation, generating highly realistic images with fine color and texture details.
Most significantly, the generated X images exhibit strong alignment with the RGB images, demonstrating its ability to generate coherent RGB+X image pairs.

Given that DiffX can generate cross-modal image pairs, we also aim to apply this framework to robust, controllable, and versatile generation across diverse modalities.
Therefore, we also conduct extensive experiments on FLIR, MFNet, and COME15K datasets for ``RGB+X+Y'' generation.
Meanwhile, we conduct experiments on MCXFace dataset for ``3DDFA $\rightarrow$ RGB+NIR+SWIR+T'' generation.
The qualitative results are shown in Fig. \ref{triple} and Fig. \ref{forth}, respectively.
It is evident that the DiffX model can effectively adapt to diverse-modal ``RGB+X+Y(+Z)'' generation under various layout guidance.
In addition, DiffX can accurately generate the images that align well with the provided text captions, where the RGB images also match the diverse-modal images effectively with the layout conditions.
Therefore, DiffX is a standout model for cohesive data augmentation across wide-range modalities.

\begin{figure}[!tbp]
  \centering
  \includegraphics[width=1\linewidth]{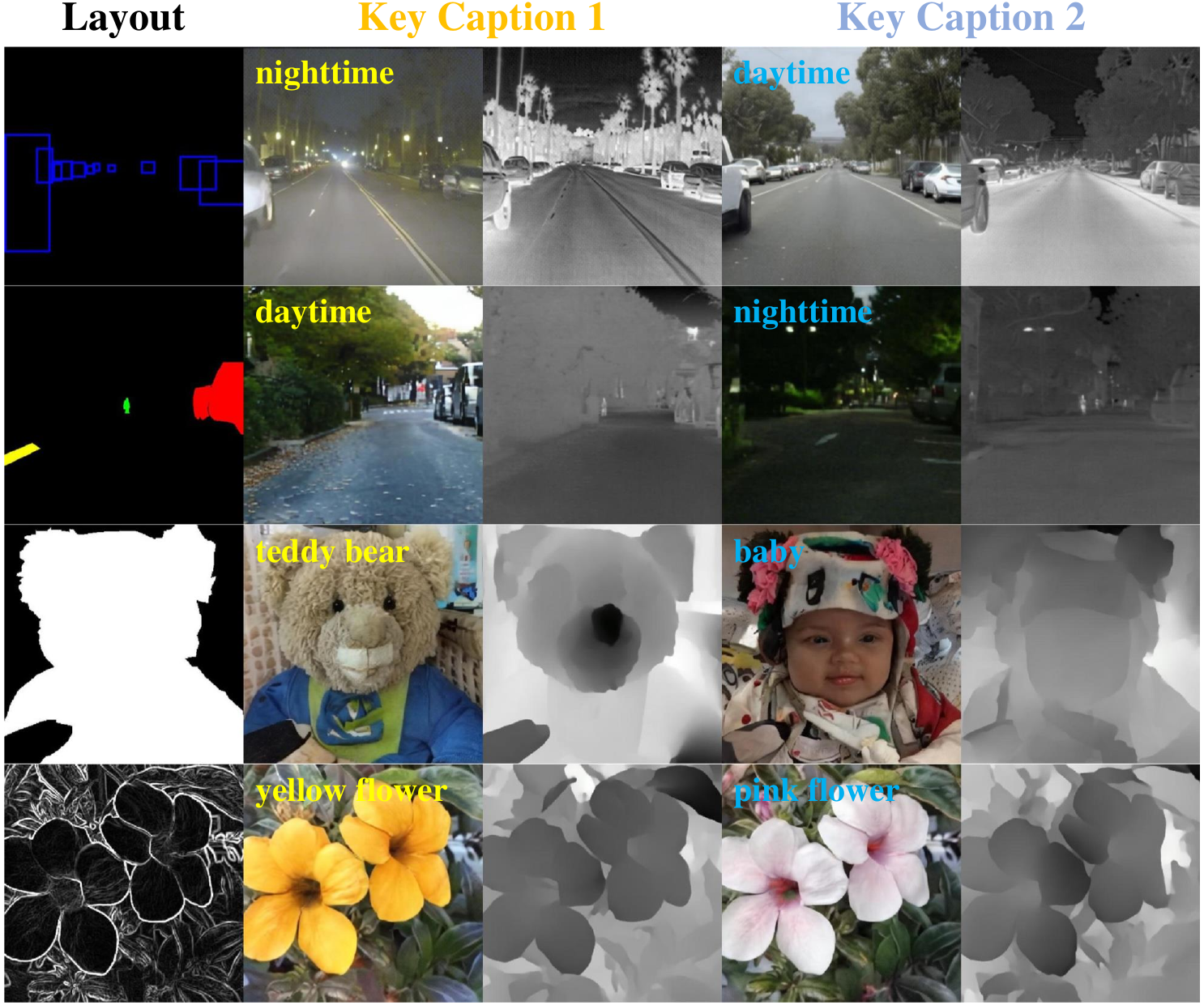}
  \caption{Diverse cross-modal generation results by editing the key captions.}
  \label{qualitative_diverse}
\end{figure}

Additionally, DiffX also enables the diverse cross-modal generation by editing the key text captions, as shown in Fig. \ref{qualitative_diverse}.
In traffic-scene RGB+T generation, the generated weather condition can be changed by modifying keywords like ``daytime'' and ``nighttime'', which is critical in autonomous driving systems.
For the natural RGB+D generation, we can change the categories and attributes of the generated objects by adjusting the captions while keeping their structural integrity and layout consistency.
Therefore, it facilitates both creative manipulation and cross-modal image editing.

\begin{figure}[!tbp]
  \centering
  \includegraphics[width=1.0\linewidth]{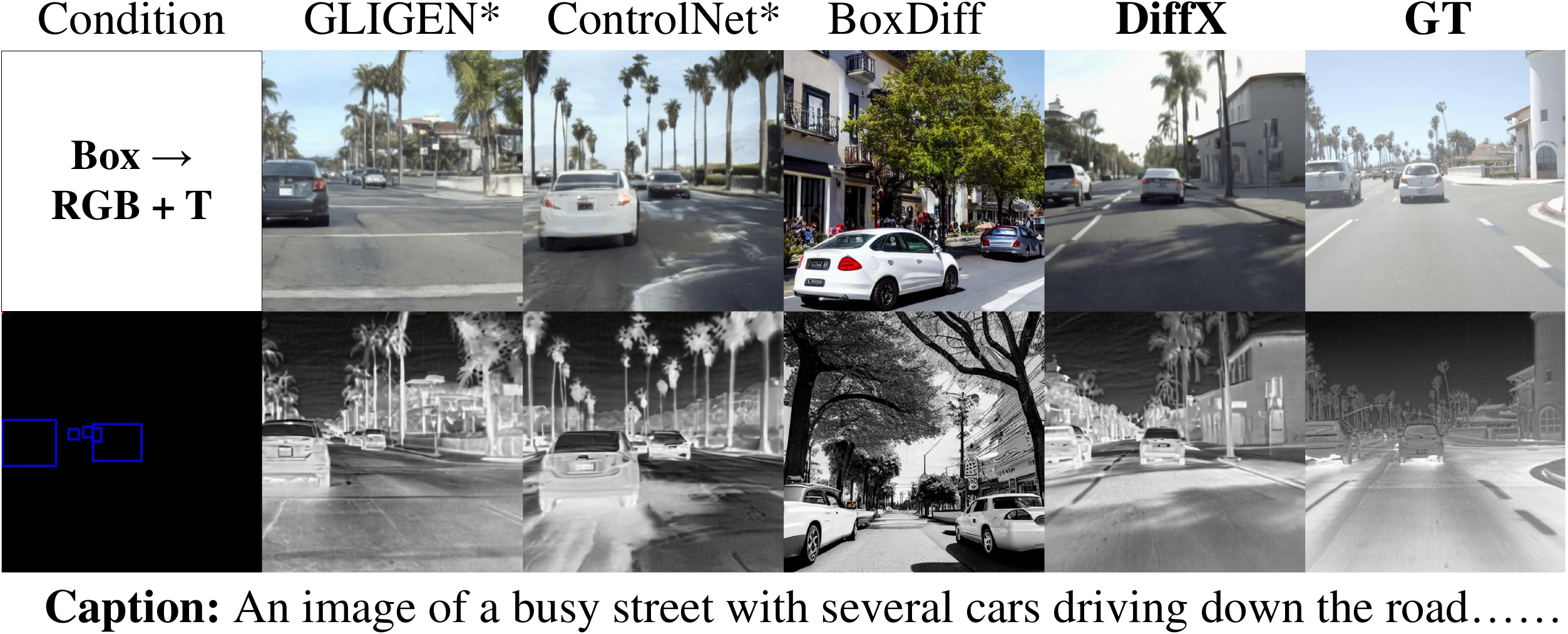}
  \caption{Qualitative comparison between our DiffX and the baseline models in task of Box $\to $ RGB+T.
  The symbol $^{*}$ denotes that the baseline models are modified to make them support cross-modal outputs.}
  \label{comparison_box}
\end{figure}

\begin{figure*}[!tbp]
  \centering
  \includegraphics[width=1.0\linewidth]{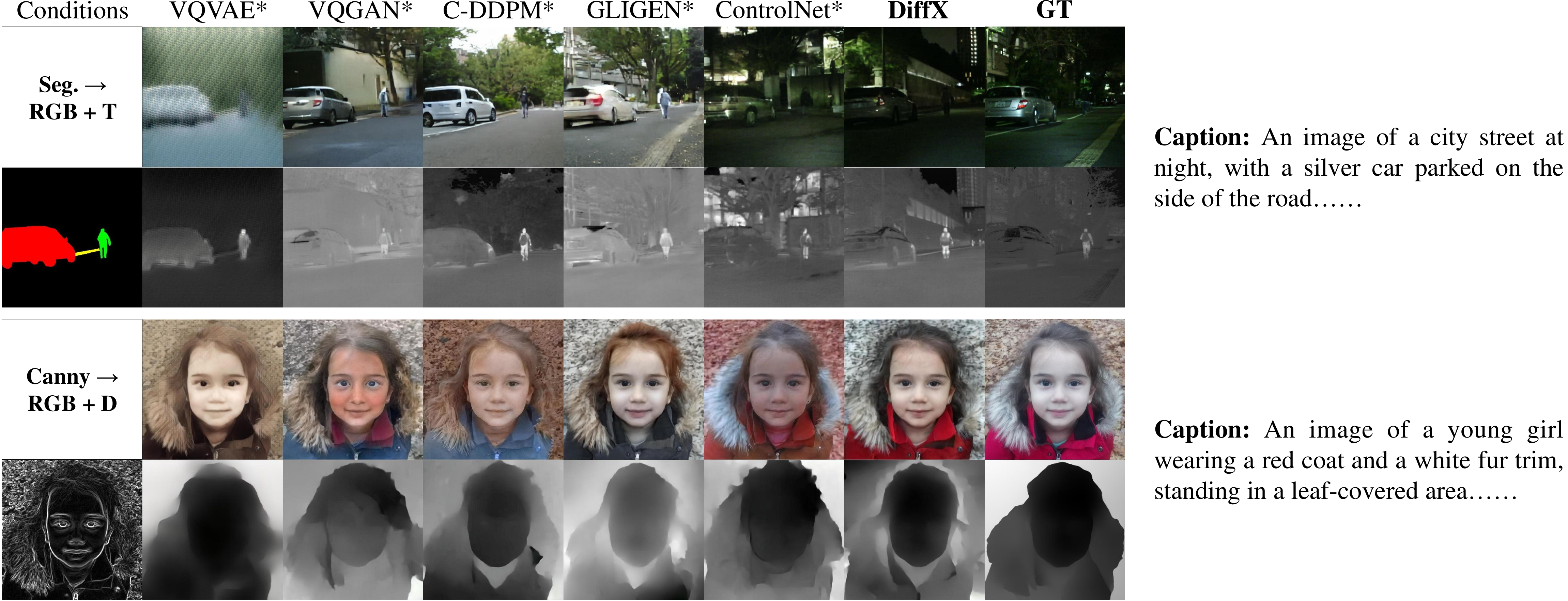}
  \caption{Qualitative comparison between our DiffX and the baseline models in tasks of Seg. $\to $ RGB+T and Canny $\to $ RGB+D.
  The symbol $^{*}$ denotes that the baseline models are modified to make them support cross-modal outputs.
  We do not include the comparison experiments in task of SOD $\to $ RGB+D because some baseline methods do not support text input and are unable to infer target object information based solely on a single SOD map.
  }
  \label{comparison}
\end{figure*}

\subsubsection{Comparison with Existing Models}
To our knowledge, our proposed DiffX is the first model for cross-modal generation.
Although the baseline models were designed for RGB image generation, we can modify them to have dual outputs for comparison.
The modifications of the specific models are described below:
In the case of end-to-end generative models such as VQVAE \cite{van2017neural} and VQGAN \cite{vqgan}, we implement them by adjusting the decoder into dual-path decoders, similar to our MP-VAE. After the modifications, we conduct the end-to-end training and testing for cross-modal ``RGB+X'' generation.
For the DDPM model \cite{ho2020denoising}, we adopt the condition incorporation method in T2V-DDPM (denoted as C-DDPM in this work) \cite{nair2023t2v} and adjust the input channel count to 3+1 by concatenating RGB and X images as the input layout conditions.
For ControlNet \cite{zhang2023adding} and GLIGEN \cite{li2023gligen}, we refine the training process by conducting diffusion in the modality-shared latent space, like our DiffX, while preserving their original structures of the denoising UNet.
For the training-free method BoxDiff \cite{xie2023boxdiff}, we keep the pre-trained model and adapt it to X image generation by editing the key text captions.

The qualitative results of our baseline models and DiffX, alongside the Ground-Truth (GT) images, are presented in Fig. \ref{comparison_box} and Fig. \ref{comparison}.
We can see that our DiffX significantly outperforms the baseline methods in task of Box $\to $ RGB+T.
While the modified GLIGEN and ControlNet can produce well-aligned RGB+T image pairs, their consistency with the layout input is subpar, primarily due to their training on datasets with large box inputs. 
BoxDiff generates relatively high-quality RGB images, but suffers from poor consistency in the image pairs due to its training-free strategy.
In tasks of Seg. $\to $ RGB+T and Canny $\to $ RGB+D, VQVAE exhibits poor generative performance due to its unsuitability for multi-path outputs in end-to-end training.
VQGAN and C-DDPM can generate overall high-quality images but lack precision in capturing target objects and details.
The modified GLIGEN, which also performs diffusion training in a modality-shared latent space, yields superior results compared with other baseline models.
However, it fails to generate accurate backgrounds in the nighttime scene despite using captions for guidance.
The modified ControlNet can generate backgrounds consistent with the provided caption, but the generated quality is worse than that of our DiffX.
Therefore, the comparison proves the effectiveness of our DiffX framework for modality-shared latent diffusion and effective joint-modal connection.

\begin{table}[!tbp]
  \setlength{\tabcolsep}{2.8pt}
  \caption{Quantitative comparison with the baseline models.}
  \centering
  \label{quantitative}
  \begin{threeparttable}
  \begin{tabular}{cc|c|cc|cc|c|c|c}
    \hline
    \multicolumn{2}{c|}{\multirow{2}{*}{Task}} & \multirow{2}{*}{Model} & \multicolumn{2}{c|}{LPIPS$\downarrow$} & \multicolumn{2}{c|}{FID$\downarrow$} & \multirow{2}{*}{AP$\uparrow$} & \multirow{2}{*}{AP$_{\text{50}}$$\uparrow$} & \multirow{2}{*}{SSIM$_x$$\uparrow$} \\
     \cdashline{4-7}
     & && RGB & X & RGB & X & & \\
    \hline
    \hline
    \multirow{4}{*}{\rotatebox{90}{Box $\rightarrow$}} & \multirow{4}{*}{\rotatebox{90}{RGB + T}} 
    & GLIGEN$^{*}$ & 0.573 & 0.538 & 67.23 & 70.23 & 12.5 & 28.6 & 0.522 \\
    && ControlNet$^{*}$ & 0.548 & 0.531 & 67.10 & 69.25 & 9.2 & 19.4 & 0.506 \\
    && BoxDiff & 0.595 & 0.583 & 71.35 & 78.94 & - & - & 0.501 \\
    && \textbf{DiffX} & \textbf{0.527} & \textbf{0.497} & \textbf{55.72} & \textbf{61.89} & \textbf{29.3} & \textbf{50.6} & \textbf{0.713}   \\
    \hline
    \hline
    \multicolumn{2}{c|}{\multirow{2}{*}{Task}} & \multirow{2}{*}{Model} & \multicolumn{2}{c|}{LPIPS$\downarrow$} & \multicolumn{2}{c|}{FID$\downarrow$} & \multirow{2}{*}{IoU$\uparrow$} & \multirow{2}{*}{Acc$\uparrow$} & \multirow{2}{*}{SSIM$_x$$\uparrow$} \\
     \cdashline{4-7}
     & && RGB & X & RGB & X & & \\
    \hline
    \hline
    \multirow{6}{*}{\rotatebox{90}{Seg. $\rightarrow$}} & \multirow{6}{*}{\rotatebox{90}{RGB + T}} 
    & VQVAE$^{*}$  & 0.673 & 0.588 & 91.58 & 98.61 & 0.437 & 0.596 & 0.508 \\
    && VQGAN$^{*}$ & 0.552 & 0.489 & 70.24 & 81.05 & 0.518 & 0.659 & 0.630 \\
    && C-DDPM$^{*}$  & 0.535 & 0.457 & 71.53 & 78.69 & 0.535 & 0.672 & 0.617 \\
    && GLIGEN$^{*}$& 0.548 & 0.428 & 69.77 & 73.58 & 0.520 & 0.667 & 0.641 \\
    && ControlNet$^{*}$& 0.541 & 0.419 & 65.32 & 73.44 & 0.512 & 0.659 & 0.638 \\
    && \textbf{DiffX} & \textbf{0.513} & \textbf{0.382} & \textbf{60.10} & \textbf{62.35} & \textbf{0.537} & \textbf{0.693} & \textbf{0.667}  \\
    \hline
    \hline
    \multicolumn{2}{c|}{\multirow{2}{*}{Task}} & \multirow{2}{*}{Model} & \multicolumn{2}{c|}{LPIPS$\downarrow$} & \multicolumn{2}{c|}{FID$\downarrow$} & \multirow{2}{*}{$S_{\alpha}$$\uparrow$} & \multirow{2}{*}{$M$$\downarrow$} & \multirow{2}{*}{SSIM$_x$$\uparrow$} \\
    \cdashline{4-7}
    & && RGB & X & RGB & X & &  \\
    \hline
    \hline
    \multirow{6}{*}{\rotatebox{90}{Canny $\rightarrow$}} & \multirow{6}{*}{\rotatebox{90}{RGB + D}} 
    & VQVAE$^{*}$  & 0.411 & 0.285 & 46.82 & 55.80  & 0.721 & 0.167 & 0.653 \\
    && VQGAN$^{*}$ & 0.389 & 0.290 & 40.82 & 41.58  & 0.733 & 0.159 & 0.747 \\
    && C-DDPM$^{*}$  & 0.255 & 0.208 & 26.79 & 34.18  & 0.751 & 0.140 & 0.768 \\
    && GLIGEN$^{*}$& 0.249 & 0.211 & 25.07 & 34.06  & 0.743 & 0.152 & 0.762 \\
    && ControlNet$^{*}$& 0.278 & 0.230 & 30.14 & 38.20 & 0.745 & 0.156 & 0.759 \\
    && \textbf{DiffX} & \textbf{0.246} & \textbf{0.197} & \textbf{17.11} & \textbf{31.20} & \textbf{0.764} & \textbf{0.134} & \textbf{0.795} \\
    \hline

  \end{tabular}
  \begin{tablenotes}
     \item $\uparrow / \downarrow$ denotes that higher / lower values are better. 
  \end{tablenotes}
  \end{threeparttable}
\end{table}

On the other hand, we evaluate the generated cross-modal image quality by the quantitative metrics, where the results are shown in Table \ref{quantitative}.
It demonstrates that our DiffX performs better than all baseline models in all metrics.
Our impressive LPIPS and FID scores demonstrate the high-quality image generation by DiffX.
For example, in the Canny $\rightarrow$ RGB + D task, DiffX achieves LPIPS scores of 0.246/0.197 and FID scores of 17.11/31.20 for RGB/X image generation, respectively, demonstrating that the quality of our generated images significantly surpasses that of the baseline models.
Moreover, the exceptional performance on downstream vision tasks highlights the alignment of generated images with the original layouts, further emphasizing the effectiveness of DiffX in data augmentation for cross-modal vision tasks.
Finally, the SSIM scores achieved on the X images illustrate the advantages of our approach for high-quality and consistent layout-aware cross-modal image generation.

\subsubsection{Comparison with Two-Stage Methods}

\begin{figure*}[!tbp]
  \centering
  \includegraphics[width=1.0\linewidth]{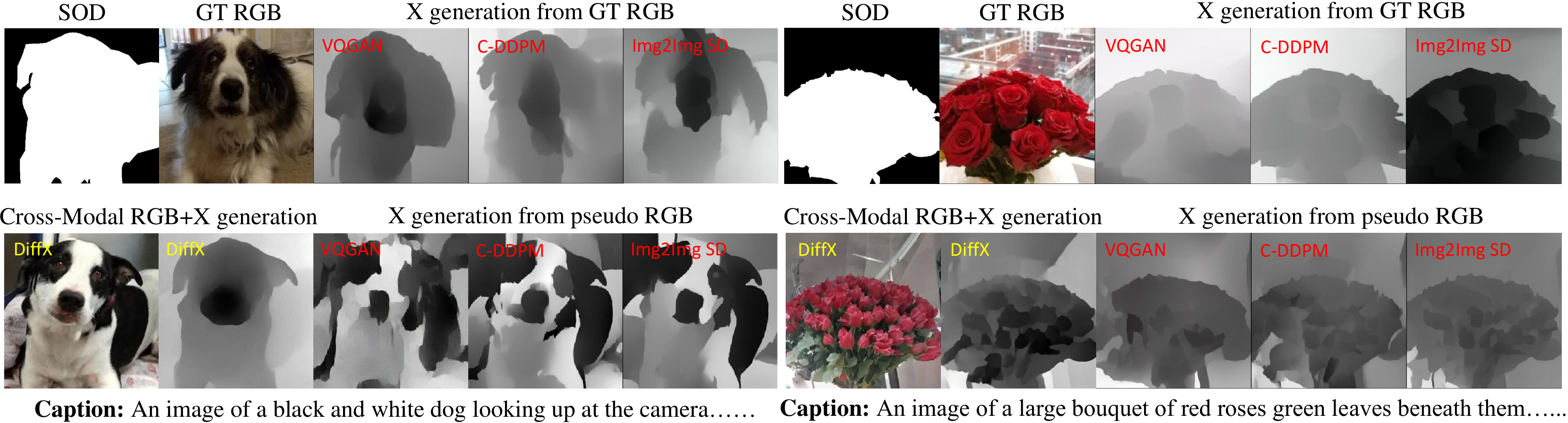}
  \caption{Qualitative comparison between our DiffX and methods of two-stage translation in the task of SOD $\rightarrow$ RGB+D generation.
  }
  \label{twostage}
\end{figure*}

\begin{table}[!tbp]
  \setlength\tabcolsep{7.0pt}
  \caption{Quantitative comparison between the uni-modal DiffX$_s$ and \textbf{cross-modal DiffX} on generated image quality.}
  \centering
  \label{lpips}
  \begin{threeparttable}
  \begin{tabular}{c|c|cc|cc}
    \hline
     \multirow{2}{*}{($\downarrow$)} & \multirow{2}{*}{Task} & \multicolumn{2}{c|}{DiffX$_s$} & \multicolumn{2}{c}{\textbf{DiffX}} \\
     \cdashline{3-6}
     & & RGB & X & RGB & X \\
    \hline
    \hline
    \multirow{4}{*}{\rotatebox{90}{LPIPS}} 
    & Box  $\rightarrow$ RGB+T & 0.557 & 0.604 & 0.527 & 0.497  \\
    & Seg. $\rightarrow$ RGB+T & 0.551 & 0.424 & 0.513 & 0.382  \\
    & SOD  $\rightarrow$ RGB+D & 0.718 & 0.489 & 0.579 & 0.305  \\
    & Canny $\rightarrow$ RGB+D & 0.258 & 0.215 & 0.246 & 0.197  \\
    \hline
    \hline
    \multirow{4}{*}{\rotatebox{90}{FID}} 
    & Box  $\rightarrow$ RGB+T & 56.31 & 76.46 & 55.72 & 61.89   \\
    & Seg. $\rightarrow$ RGB+T & 71.58 & 75.17 & 60.10 & 62.35   \\
    & SOD  $\rightarrow$ RGB+D & 71.23 & 80.54 & 51.59 & 52.83   \\
    & Canny $\rightarrow$ RGB+D & 19.36 & 35.60 & 17.11 & 31.20   \\
    \hline
  \end{tabular}
  \end{threeparttable}
\end{table}

DiffX employs an integrated diffusion model to simultaneously generate cross-modal images in one stage.
Traditional models that generate RGB and X images separately often result in misaligned and inconsistent image pairs.
Alternatively, a generative model could first produce RGB images, followed by another model generating X images based on the RGB outputs to achieve RGB+X generation.
However, this two-stage translation approach tends to yield poor generative fitting for the second-stage generation.
To illustrate this, we conduct experiments using VQGAN \cite{vqgan}, C-DDPM \cite{ho2020denoising}, and Img2Img SD \cite{rombach2022high} for Layout $\rightarrow$ RGB $\rightarrow$ X translation, namely the two-stage translation, and the results are shown in Fig. \ref{twostage}.
The first row depicts the X images generated from Ground-Truth (GT) RGB images, while the second row displays cross-modal results generated by DiffX alongside X results derived from pseudo RGB images (the first column).
The results indicate that although conventional models can infer X images from GT RGB images, the pixel-level inconsistency still exists.
Moreover, the X images generated from pseudo RGB images exhibit poor quality due to the models being trained on GT RGB $\rightarrow$ X translation but inferred on generated RGB images.
Consequently, the two-stage translation using conventional generative models fails to ensure consistent and high-quality cross-modal generation, whereas our DiffX maintains superior efficiency and performance.

\subsubsection{Ablation Study}
\begin{table}[!tbp]
  \setlength\tabcolsep{5.0pt}
  \caption{Ablation study on the Laplacian Pyramid (LP).}
  \centering
  \label{laplacian}
  \begin{threeparttable}
  \begin{tabular}{c|c|c|cc|cc}
    \hline
    \multirow{2}{*}{($\uparrow$)} & \multirow{2}{*}{Dataset} & \multirow{2}{*}{Source} & \multicolumn{2}{c|}{MP-VAE w/o LP} & \multicolumn{2}{c}{\textbf{MP-VAE w/ LP}} \\
     \cdashline{4-7}
     & & & RGB & X & RGB & X \\
    \hline
    \hline
    \multirow{3}{*}{\rotatebox{90}{PSNR}} 
    & FLIR & RGB+T   & 29.04 & 29.77 & 29.25 & 30.89  \\
    & MFNet& RGB+T   & 30.59 & 30.96 & 31.45 & 32.86  \\
    & COME & RGB+D   & 30.09 & 34.19 & 30.15 & 35.73  \\
    \hline
    \hline
    \multirow{3}{*}{\rotatebox{90}{SSIM}} 
    & FLIR & RGB+T   & 0.785 & 0.719 & 0.792 & 0.737  \\
    & MFNet& RGB+T   & 0.788 & 0.845 & 0.800 & 0.919  \\
    & COME & RGB+D   & 0.729 & 0.958 & 0.749 & 0.974  \\
    \hline
  \end{tabular}
  \begin{tablenotes}
     \item w/ and w/o denote with and without, respectively. 
  \end{tablenotes}
  \end{threeparttable}
\end{table}

To verify the effectiveness of our proposed strategies for cross-modal generation, we conduct the ablation study on the different modules in the DiffX model.

Firstly, we aim to compare cross-modal RGB+X generation by DiffX with uni-modal generation by DiffX variant.
Therefore, we convert DiffX into a standard generative model (denoted as DiffX$_s$), which does not adopt the modality-shared latent diffusion, to generate RGB and X images separately.
The quantitative comparison in Table \ref{lpips} shows that the cross-modal results generated by our proposed DiffX exhibit better LPIPS and FID scores.
Notably, in the SOD  $\rightarrow$ RGB+D task, DiffX achieves FID scores of 51.59 and 52.83 for RGB and X image generation, respectively, while the DiffX$_s$ variant records the scores of 71.23 and 80.54.
This improvement can be attributed to the diffusion and denoising processes conducted in a modality-shared latent space, which provides complementary information for the joint RGB+X image generation.
Therefore, the proposed modality-shared mechanism is essential for cross-modal generation, especially in the pixel-aligned image pairs or groups.

Secondly, we conduct the quantitative ablation study on the adopted Laplacian Pyramid (LP) in our MP-VAE, where we compare the reconstruction results of MP-VAE and that w/o LP.
Results in Table \ref{laplacian} show that the LP structure effectively improves the PSNR and SSIM scores on FLIR, MFNet, and COME datasets for RGB+X reconstruction, facilitating cross-modal reconstruction through frequency-aware information incorporation.
For example, on the MFNet dataset, the MP-VAE w/ LP achieves the PSNR of 31.45 and 32.86 for the reconstruction of RGB and X images, outperforming the MP-VAE w/o LP by 0.86 and 1.90, respectively.

\begin{figure}[!tbp]
  \centering
  \includegraphics[width=1\linewidth]{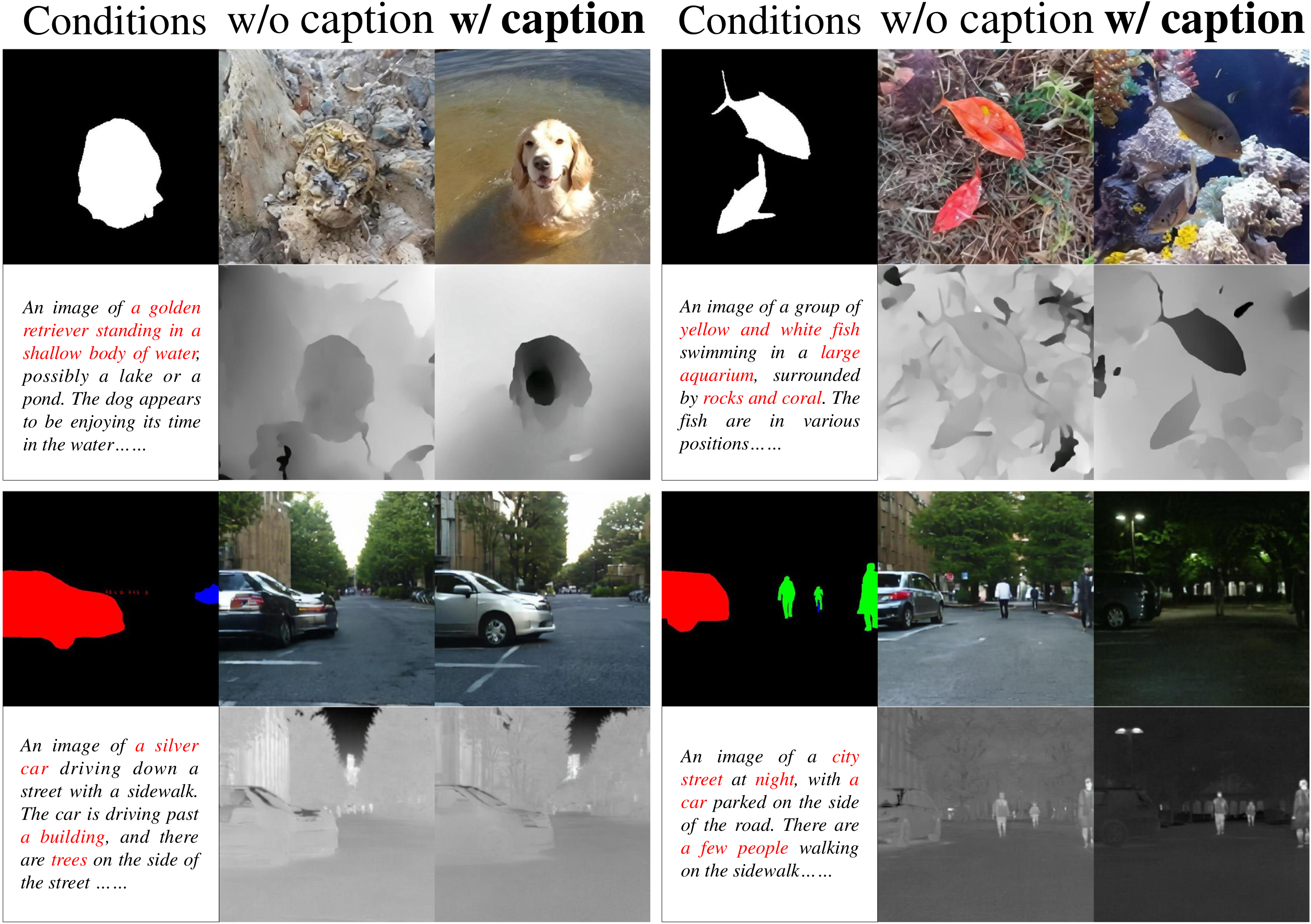}
  \caption{Impact of embedding text captions in cross-modal generation.
  The variant models without using caption embeddings generate broken or misaligned images, ultimately affecting the image quality.
    }
  \label{SOD_caption}
\end{figure}

Thirdly, we conduct the ablation study on the impact of long text captions on SOD $\rightarrow$ RGB+D and Seg. $\rightarrow$ RGB+T tasks.
The qualitative comparison in Fig. \ref{SOD_caption} shows that DiffX can effectively capture the crucial captions, while the model without caption embeddings generates broken or misaligned images, ultimately affecting the image quality.
Therefore, it also necessitates incorporating the JME module for long text embedding and joint-modality connection.

Finally, it is essential to validate the effectiveness of our proposed JME module.
We conduct experiments for the task of Canny $\rightarrow$ RGB+D, and the results are shown in Table \ref{ablation_wo}.
The variant w/o GC\&FF removes the gated cross-attention and FF layers, directly feeding $\boldsymbol{c}^{*}$ and $\boldsymbol{H}$ into the DiffX-UNet.
The variant labeled w/o gate eliminates the gated mechanism in the gated cross-attention layer.
From the results, we can see that the JME module still plays a significant role in condition injection, as DiffX shows improvement across most metrics.
There is a coincidence that the variant w/o gate achieves the LPIPS of 0.195, which is lower than that of DiffX by 0.002.

\begin{table}[!tbp]
  \setlength{\tabcolsep}{4pt}
  \caption{Ablation study on the JME module in task of Canny $\rightarrow$ RGB+D.}
  \centering
  \label{ablation_wo}
  \begin{threeparttable}
  \begin{tabular}{c|cc|cc|c|c|c}
    \hline
    \multirow{2}{*}{Model} & \multicolumn{2}{c|}{LPIPS$\downarrow$} & \multicolumn{2}{c|}{FID$\downarrow$} & \multirow{2}{*}{$S_{\alpha}$$\uparrow$} & \multirow{2}{*}{$M$$\downarrow$} & \multirow{2}{*}{SSIM$_x$$\uparrow$} \\
    \cdashline{2-5}
    & RGB & X & RGB & X & &  \\
    \hline
    \hline
    \textbf{DiffX} & \textbf{0.246} & 0.197 & \textbf{17.11} & \textbf{31.20} & \textbf{0.764} & \textbf{0.134} & \textbf{0.795} \\
    w/o GC\&FF & 0.252 & 0.200 & 20.87 & 33.56 & 0.755 & 0.143 & 0.774 \\
    w/o gate & 0.249 & \textbf{0.195} & 19.24 & 33.28 & 0.760 & 0.138 & 0.776 \\
    \hline
  \end{tabular}
  \begin{tablenotes}
     \item $\uparrow / \downarrow$ denotes that higher / lower values are better. 
  \end{tablenotes}
  \end{threeparttable}
\end{table}

\section{Discussion}

\subsection{Advantages over Existing Diffusion Models}
Through the extensive experiments, our joint-learning mechanism in modality-shared latent space has proven to be highly effective.
From the experimental analysis, it is evident that DiffX outperforms both two-stage translation models and other multi-modal generative variants.
This performance aligns with our expectations and demonstrates DiffX's ability to sample more robust distributions by fully leveraging the multi-modal data.
In addition, handling multimodal data inherently introduces additional computational overhead. However, in our joint-modality setting, this overhead is minimal.
DiffX's multi-path decoders are lightweight, and optimizing for long text captions does not significantly increase computation.
Compared to the classic single-modal layout-to-image model GLIGEN, DiffX requires less than only 1\% more inference time.
While there is a slight increase in computational demand for multi-path decoding, our method remains more efficient in terms of both time and storage compared to training two separate models for each modality.
Furthermore, compared with the widely-used model ControlNet, DiffX can support integrated layout guidance and flexible output formats (such as the modalities besides RGB images).

\subsection{Current Limitations}
The proposed DiffX method embeds cross-modal images into a modality-shared latent space.
During the training process, DiffX places stringent demands on the quality of training data, necessitating pixel-aligned image pairs or groups.
Although we have endeavored to collect a comprehensive array of cross-modal datasets and implemented various strategies to supplement the required image modalities, the availability of public data of this type is limited.
Furthermore, the existing datasets often suffer from low image quality, singular scene representation, and inadequate generalization capabilities.
For instance, the RGB+T datasets utilized in this study, specifically the FLIR and MFNet datasets, are both derived from road scenes.
This presents challenges when generating RGB+T image pairs for other contexts, such as indoor or industrial environments, as the trained model exhibits limited generalization abilities.
Consequently, compared to many contemporary diffusion models that generate images from texts, the primary obstacle in the domain of cross-modal image generation lies in the scarcity of high-quality and diverse cross-modal image data with accurate layout annotations.

\subsection{Future Prospects}
On one hand, as previously highlighted, a key limitation in cross-modal generation stems from the requirement for cross-modal data.
Consequently, we aim to broaden the dataset scope and enhance the diversity and unity of the modalities to facilitate the training of a more robust and generalized DiffX.

On the other hand, our cross-modal generative framework demonstrates flexibility and is well-suited for various practical applications.
As the first framework to leverage diffusion methodology for layout-guided cross-modal generation tasks, we have emphasized multiple modalities to showcase DiffX's effectiveness and potential.
We present diverse modal generation results on the COME15K and MCXFace datasets, paving the way for future research that incorporates additional modalities.
This expansion promises to enhance the versatility and applicability of DiffX in a range of real-world scenarios.
For example, by integrating viewing angle information into the RGB+D generation process, we can directly produce 3D images from SOD maps, thereby improving depth and realism. 
Furthermore, if we develop a robust cross-modal audio-video encoding method, we would be able to generate corresponding audio and video content based on user requirements, further broadening the framework's versatility and applicability.

\section{Conclusion}
In this study, we introduce the novel diffusion model, named DiffX, for cross-modal ``RGB+X(+Y+Z)'' image generation under various layout conditions and text captions.
The key innovation lies in its novel diffusion-based generative pipeline within the modality-shared latent space, where each modality functions independently.
Extensive experiments have demonstrated its robustness and adaptability in cross-modal generation.
Meanwhile, the ablation study illustrates the effectiveness of the proposed modules and strategies for joint-modality learning.
Moreover, it maintains superior efficiency when compared to conventional generative models and two-stage translation methods in cross-modal generation.
Moving forward, we aim to enhance our DiffX by enabling multiple-to-multiple generative modeling, allowing for the generation of unrestricted modalities.
Additionally, we also encourage more researchers to focus on the field of cross-modal generation and collaborate in creating large amounts of high-quality training data to advance the development of this area.

\bibliographystyle{IEEEtran}
\bibliography{citation.bib}

\vspace{11pt}

\vfill

\end{document}